 \documentclass[smallabstract,smallcaptions]{dccpaper}

\usepackage{epsfig}
\usepackage{amsmath}
\usepackage{amssymb}
\usepackage{color}
\usepackage{url}
\usepackage{pifont}

\usepackage{algorithm}
\usepackage{algpseudocode}  

\usepackage{booktabs}
\usepackage{multirow}
\usepackage{tabularx}
\usepackage{makecell}
\usepackage{balance} 
\usepackage{microtype} 
\usepackage{caption}
\newcommand{\cmark}{\checkmark}
\newcommand{\xmark}{$\times$}

\newlength{\figurewidth}
\newlength{\smallfigurewidth}

\begin{document}

\title{\large
\textbf{Content Adaptive based Motion Alignment Framework for Learned Video Compression}
}

\author{%
Tiange Zhang$^{1,2}$, Xiandong Meng$^{2,\ast}$ and Siwei Ma$^{1,2,\ast}$\thanks{$^{\ast}$Corresponding authors: Siwei Ma (\protect\url{swma@pku.edu.cn}) and Xiandong Meng (\protect\url{mengxd@pcl.ac.cn}). This work was supported in part by the Key Research \& Development Program of Peng Cheng Laboratory under grant PCL2024A02.}\\[0.5em]
{\small\begin{minipage}{\linewidth}\begin{center}
\begin{tabular}{c}  
$^{1}$ Shenzhen Graduate School, Peking University, Shenzhen, China \\
$^{2}$Pengcheng Laboratory, Shenzhen, China
\end{tabular}
\end{center}\end{minipage}}
}

\maketitle
\thispagestyle{empty}

\begin{abstract}
Recent advances in end-to-end video compression have shown promising results owing to their unified end-to-end learning optimization. However, such generalized frameworks often lack content-specific adaptation, leading to suboptimal compression performance. To address this, this paper proposes a content adaptive based motion alignment framework that improves performance by adapting encoding strategies to diverse content characteristics. Specifically, we first introduce a two-stage flow-guided deformable warping mechanism that refines motion compensation with coarse-to-fine offset prediction and mask modulation, enabling precise feature alignment. Second, we propose a multi-reference quality aware strategy that adjusts distortion weights based on reference quality, and applies it to hierarchical training to reduce error propagation. Third, we integrate a training-free module that downsamples frames by motion magnitude and resolution to obtain smooth motion estimation. Experimental results on standard test datasets demonstrate that our framework CAMA achieves significant improvements over state-of-the-art Neural Video Compression models, achieving a 24.95\% BD-rate (PSNR) savings over our baseline model DCVC-TCM, while also outperforming reproduced DCVC-DC and traditional codec HM-16.25.
\end{abstract}

\Section{Introduction}
Recent years have witnessed remarkable progress in end-to-end learned video compression, where frameworks based on deep networks have achieved continuous breakthroughs, surpassing traditional codecs such as H.264/AVC~\cite{wiegand2003overview}, H.265/HEVC~\cite{sullivan2012overview} and H.266/VVC~\cite{bross2021overview} in rate-distortion (RD) performance. Unlike traditional methods that require manual tuning and separate optimization of individual modules, end-to-end learning enables joint optimization of all components through unified training, directly targeting overall RD performance optimization.

Among various end-to-end video compression frameworks~\cite{jia2025emerging,SIP-20250056,10222075,LIN2023126396,liao2025ehvc}, conditional coding paradigm has emerged as a dominant approach due to its superior RD performance. This paradigm can be traced back to DVC~\cite{lu2019dvc}, which replaced handcrafted modules with fully trainable networks in the pixel domain. FVC~\cite{hu2021fvc} further introduced feature domain modeling for more compact representations. The introduction of DCVC~\cite{li2021deep} established the first conditional coding framework, where both encoder and decoder are conditioned on features extracted from previously reconstructed frames to better exploit temporal dependencies.

Following the conditional coding paradigm, recent works have continuously advanced neural video compression along three major directions: motion modeling, context modeling, and entropy modeling. In motion modeling, recent works mainly focus on refining motion representation such as decomposing motion into different components, employing bi-directional motion reference to capture complex dynamics. In context modeling, recent efforts focus on capturing multi-scale temporal contexts and enhancing context diversity across both temporal and spatial dimensions to better exploit correlations. In entropy modeling, recent approaches aim at leveraging both temporal and spatial priors to improve the accuracy of probability estimation.

Despite these advances, existing neural video compression (NVC) methods still struggle with content-adaptive motion alignment—the ability to dynamically adjust motion modeling according to scene characteristics. Many recent approaches rely on deformable alignment and explicitly transmit learned offsets~\cite{zhai2024hybrid,yilmaz2023multi,hu2022coarse}, which improves motion alignment but inevitably increases bitrate consumption. Moreover, motion models trained under fixed hierarchical quality structures, such as DCVC-DC~\cite{li2023neural}, cannot flexibly adapt to temporal content variations, leading to error accumulation. In addition, most flow estimation modules overlook differences in resolution and motion magnitude, resulting in unstable alignment, particularly for high-resolution or fast-motion sequences. While some methods~\cite{tang2024offline,lu2020content} employ online training to mitigate such issues, they impose severe overhead on encoding time.

To address these challenges, we propose a Content-Adaptive Motion Alignment (CAMA) framework that explicitly adapts motion modeling and flow estimation to the visual characteristics and dynamics of each sequence. Our contributions are summarized as follows:
\noindent\begin{itemize}
\setlength{\leftskip}{-0.4em} \item We introduce a two-stage motion compensation module, which adopts flow-guided deformable alignment to refine motion compensation and achieve precise alignment through coarse-to-fine offsets and masks in multiple scales.
\item We propose a multi-reference quality aware hierarchical training structure, which achieves better quality fluctuations and reduces error propagation by perceiving the quality changes of inter-frame motion.
\item We effectively integrate a training-free smooth motion estimation method that adaptively downsamples inputs according to motion magnitude and resolution, enabling more stable and accurate flow in complex motion scenarios.
\item Experimental results show that our proposed method notably improves upon the baseline model DCVC-TCM, achieving 24.95\% BD-Rate gain in terms of PSNR. It also outperforms SOTA DCVC-DC and traditional codec HM-16.25.
\end{itemize}

\Section{Proposed Method}
\subsection*{\textbf{Overall Framework}}

\begin{figure*}[t]
\centering
\includegraphics[width=0.9\textwidth]{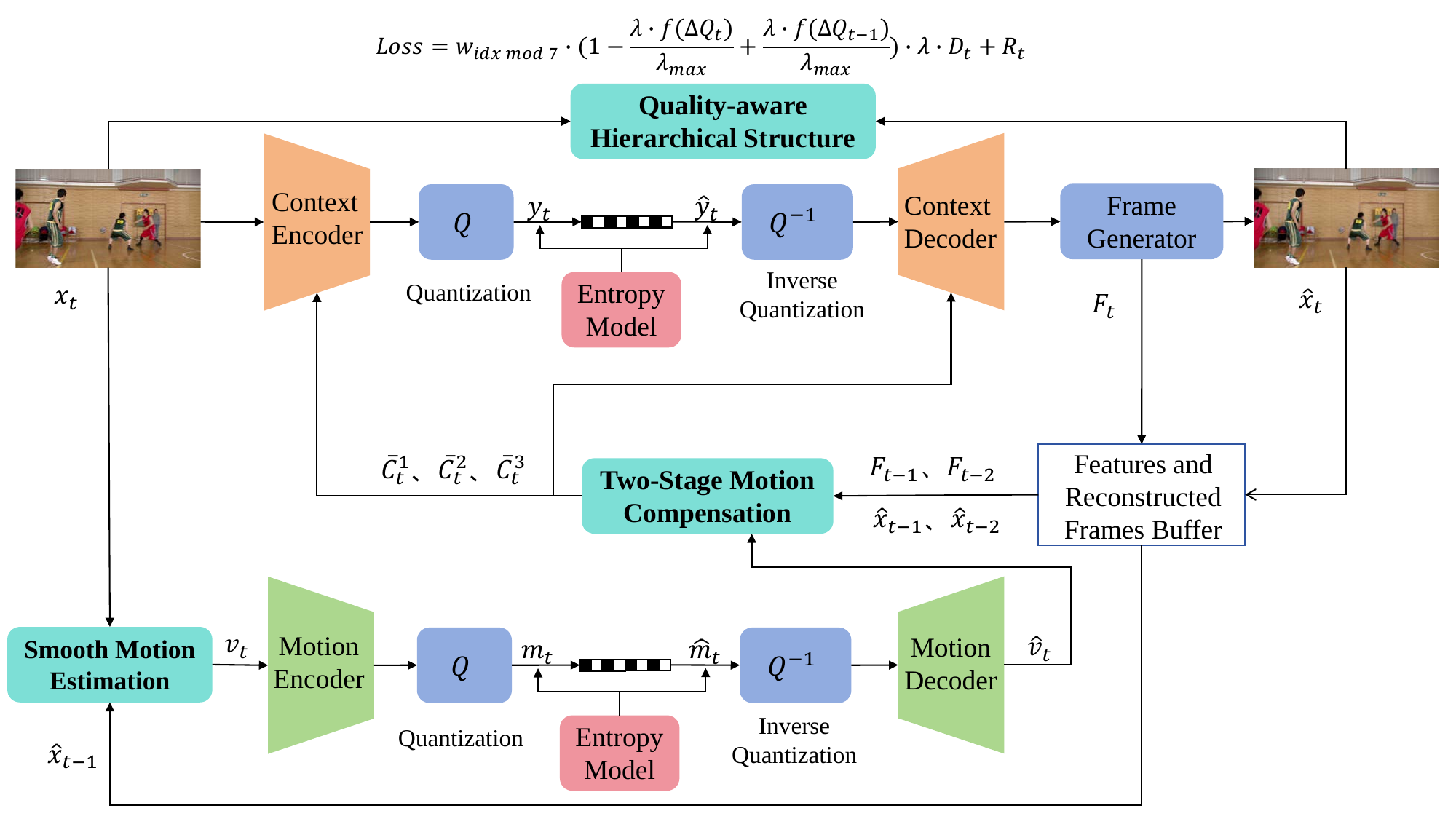}
\caption{Overall architecture of our proposed CAMA framework.}
\label{figs:framework}
\vspace{-1em}
\end{figure*}

The CAMA framework extends the DCVC-TCM~\cite{sheng2022temporal} motion-context encoder-decoder, incorporating DCVC-DC~\cite{li2023neural} improvements like deep separable networks. It enhances motion alignment with Smooth Motion Estimation for precise flow, Two-Stage Motion Compensation to refine alignment at each scale, and a Quality-aware Hierarchical Structure for adaptive training based on motion quality, as shown in Fig.~\ref{figs:framework}. This optimizes motion alignment accuracy and leverages scene characteristics with a hierarchical structure sensitive to quality variations.

For an input frame $x_t$, we estimate smooth optical flow $v_t$ between $x_t$ and $\hat{x}_{t-1}$, which is then quantized, and entropy encoded. The flow $\hat{v}_t$ and reference features are downsampled to multi-scale representations $\hat{v}_t^s$ and $F_{t-1}^s$ ($s \in \{1,2,3\}$). Two-Stage Motion Compensation refines alignment by fusing $F_{t-1}^s$ and $\hat{v}_t^s$ to predict offsets and modulation masks, guiding deformable convolution to produce aligned features $C_t^s$, which are further refined by the context module to yield $\bar{C}_t^s$.

Context encoding processes $x_t$ with $\bar{C}_t^s$ to produce $y_t$, which is quantized, passed to the entropy model, and decoded symmetrically. The Multi-Reference Quality Aware Hierarchical Structure computes a multiplier term to adjust the frame’s quality weight $w_t$, reducing error propagation. Finally, $\hat{y}_t$ recovers the final frame $\hat{x}_t$ under $\bar{C}_t^s$, with $\hat{x}_t$ stored for future coding. Detailed module structures are in the following sections.

\subsection*{\textbf{Two-Stage Motion Compensation with Flow-Guided Deformable Warp}}

\begin{figure}[t]
\centering
\includegraphics[width=0.7\textwidth]{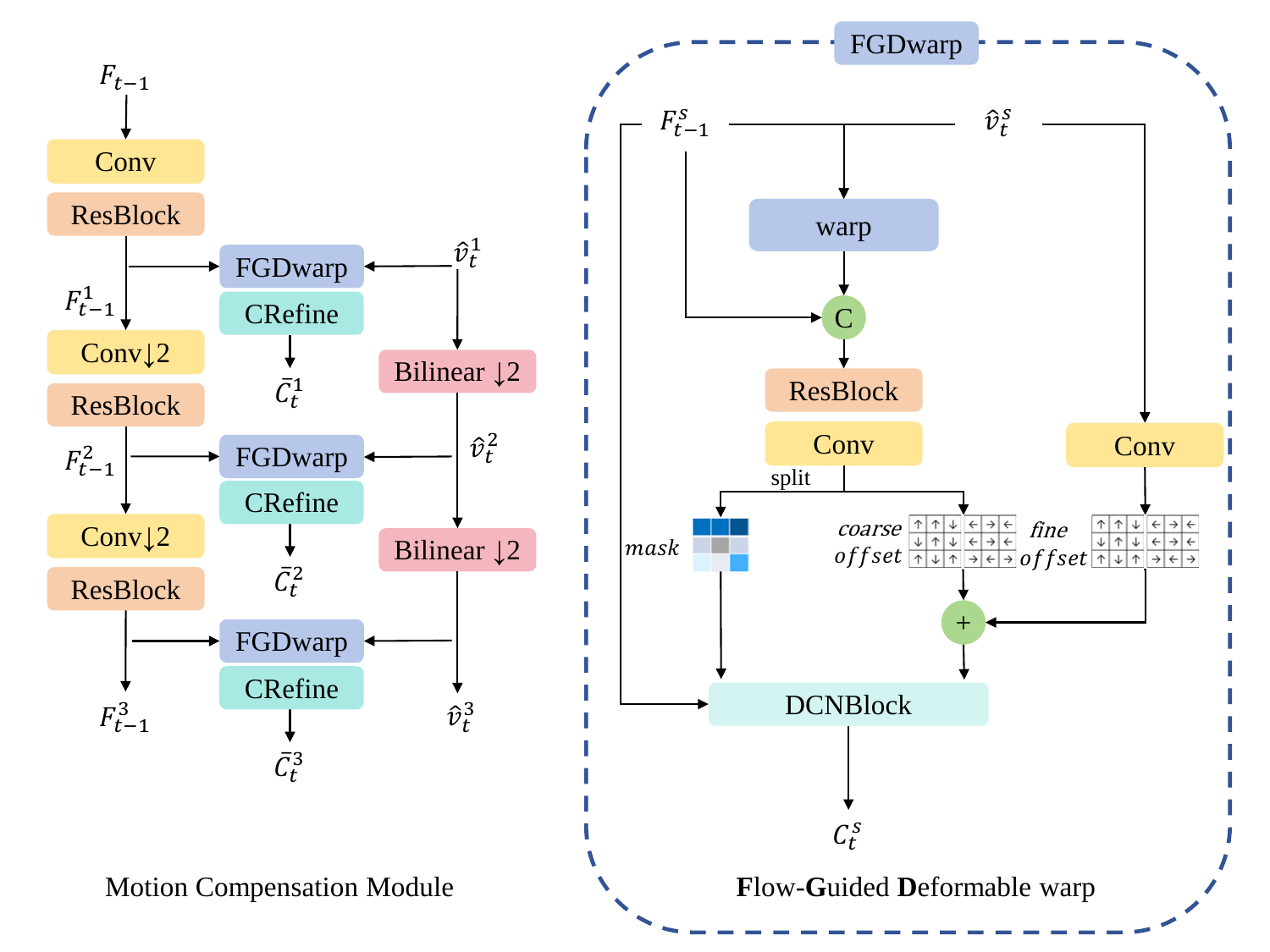}
\caption{The structure of the two-stage motion compensation module.}
\label{figs:detail}
\vspace{-1em}
\end{figure}

Motion compensation is vital in learned video compression as it provides conditional information for prediction. Existing methods typically warp reference feature $F_{t-1}$ using reconstructed flow $\hat{v}_t$, which works for simple motion but produces large residuals under fast or non-rigid motion. Some approaches introduce deformable alignment with transmitted offsets to refine alignment, but this inevitably increases bitrate.

We observe that $\hat{v}_t$ already contains sufficient local displacement cues to infer fine-grained offsets without side information. Thus, we design a Two-Stage Motion Compensation(TSMC) scheme: a coarse flow-guided warp followed by a Flow-Guided Deformable Warp (FGDwarp) module, which predicts offsets directly from $\hat{v}_t$ and $F_{t-1}$ for more accurate motion alignment. The refined features are then used for subsequent compression.

\textbf{Coarse-to-fine multi-scale offset and mask prediction.}
As shown in Fig.~\ref{figs:detail}, FGDwarp adopts a coarse-to-fine multi-scale structure with three scale levels $s \in \{1, 2, 3\}$. We extract multi-scale features $\{F_{t-1}^1, F_{t-1}^2, F_{t-1}^3\}$ from the previous reconstructed frame $\hat{x}_{t-1}$ by applying strided convolutions and resblocks. The optical flow $v_t$ is similarly downsampled via bilinear interpolation to obtain $\{v_t^1, v_t^2, v_t^3\}$.For each scale, we first apply  bilinear warp to align the reference feature $F_{t-1}^s$ using the flow $\hat{v}_t^s$:
\begin{equation}
\tilde{F}_{t-1}^s = \text{Warp}(F_{t-1}^s, \hat{v}_t^s).
\end{equation}

The warped feature $\tilde{F}_{t-1}^s$ is then concatenated with the reconstructed frame feature $F_{t-1}^s$ and passed through a residual block followed by a convolutional layer to extract a hidden representation:
\begin{equation}
H_t^s = \text{Conv}(\text{ResBlock}(\text{concat}[\tilde{F}_{t-1}^s, F_{t-1}^s])).
\end{equation}

This intermediate feature $H_t^s$ is split into two parts along the channel dimension, one for predicting a coarse sampling offset $O_{\text{coarse}}^s$, and the other for predicting the modulation mask $M_t^s$:
\begin{align}
O_{\text{coarse}}^s &= \text{Split}_{\text{offset}}(H_t^s),\\
M_t^s &= f(\text{Split}_{\text{mask}}(H_t^s)),
\end{align}
where $f(\cdot)$ denotes the sigmoid function applied to normalize the mask to $[0, 1]$.

Meanwhile, the flow $\hat{v}_t^s$ is also passed through an independent convolutional branch to estimate a fine-grained offset:
\begin{equation}
O_{\text{fine}}^s = \text{Conv}(\hat{v}_t^s).
\end{equation}

The final sampling offset $O_t^s$ is obtained by summing the coarse and fine predictions:
\begin{equation}
O_t^s = O_{\text{coarse}}^s + O_{\text{fine}}^s.
\end{equation}

\textbf{Deformable feature alignment.}
At last, we use deformable convolution to align the original reference feature $F_{t-1}^s$ with the predicted offset and mask:
\begin{equation}
C_t^s = \text{DCNBlock}(F_{t-1}^s, O_t^s, M_t^s),
\end{equation}
where $C_t^s$ is the aligned context feature at scale $s$. The three-level features $\{C_t^1, C_t^2, C_t^3\}$ are then forwarded into the context refine module made up of resblocks for further refinement.

\begin{figure}[t]
\centering
\includegraphics[width=0.8\textwidth]{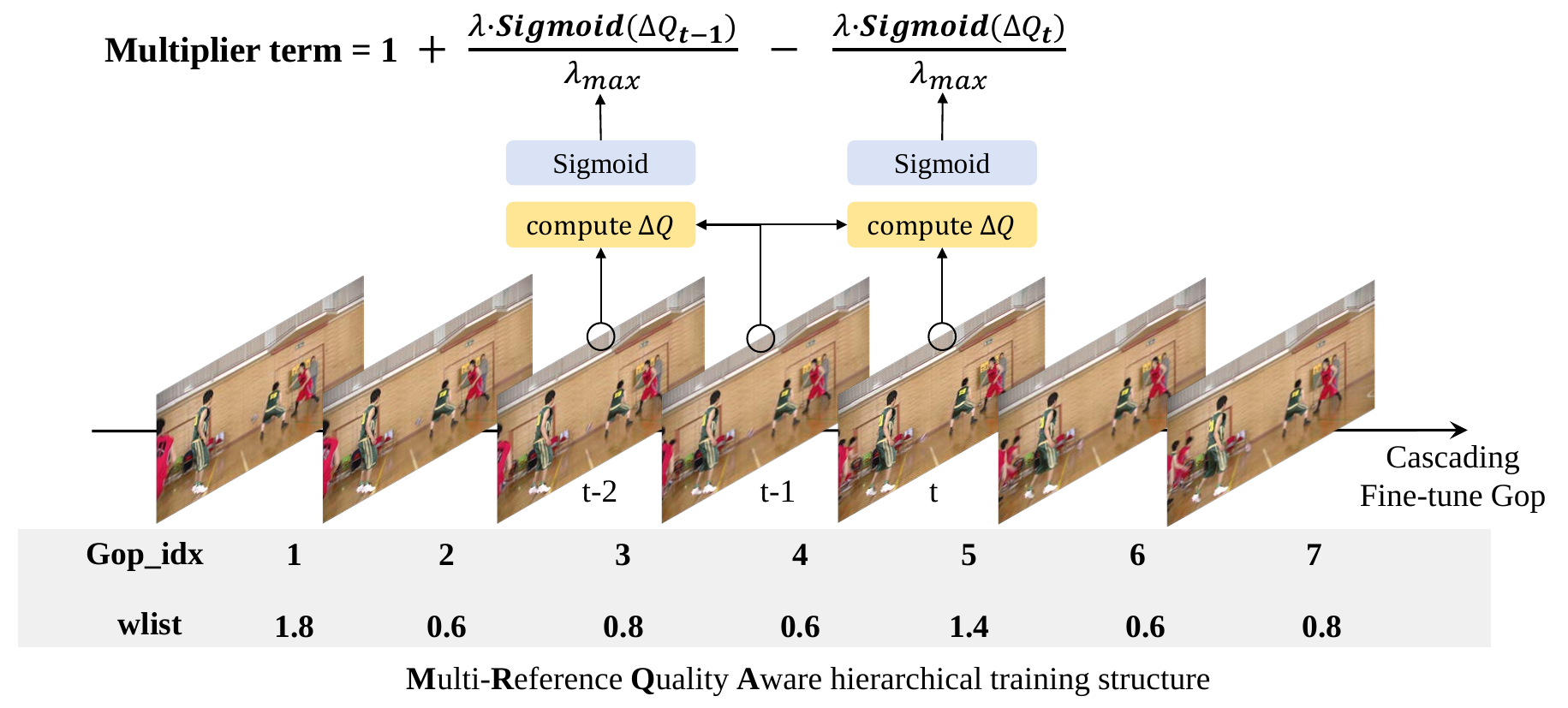}
\caption{The strategy of multi-reference quality aware hierarchical
structure training.}
\label{figs:MRQA}
\vspace{-1em}
\end{figure}

\subsection*{\textbf{Multi-Reference Quality Aware Hierarchical Training Structure}}
Existing works in neural video compression have demonstrated that periodically enhancing the quality of reference frames can yield significant Rate-Distortion (RD) gains. Motivated by this, hierarchical training strategies, such as the ``three-low-one-high'' pattern (e.g., weights like $[1.8, 0.6, 0.6, 0.6]$), were proposed to simulate such temporal variations. By assigning higher weights to specific frames within a mini-GOP, these methods aim to provide high-fidelity references for subsequent predictions. However, we identify that this fixed ``three-low-one-high'' strategy is suboptimal. Its rigid setting relies on hand-crafted hyperparameters that fail to adapt to actual content characteristics, limiting further performance improvements.

In practice, content complexity and motion intensity fluctuate dynamically, causing fixed-weight training to mismatch the actual importance of references. For instance, when a frame assigned a low weight unexpectedly serves as a critical reference for complex motion, its quality degradation propagates to subsequent frames, leading to temporal inconsistency and reduced reconstruction fidelity.

To mitigate this, we propose a Multi-Reference Quality Aware (MRQA) strategy that adaptively modulates the training weight $w_t$ of each frame according to temporal quality variation. MRQA monitors the reconstructed quality fluctuation to infer reference stability, and accordingly adjusts hierarchical weights during training. Following ~\cite{zuo2024frame}, we extend the dependency window from 4 to 7 frames with empirically set weights $[1.8, 0.6, 0.8, 0.6, 1.4, 0.6, 0.8]$, which reduce the extreme weight difference by about 0.2. This extension achieves smoother quality transitions and reduced temporal quality fluctuation. Specifically, we first compute the frame-level PSNR variation:
\begin{equation}
\Delta Q_t = \mathrm{PSNR}_{\hat{x}_t} - \mathrm{PSNR}_{\hat{x}_{t-1}} .
\end{equation}

This value reflects the relative quality change between adjacent frames. Then, a sigmoid activation is applied to normalize the quality change:
\begin{equation}
f(\Delta Q_t) = \text{sigmoid}(\Delta Q_t).
\end{equation}

The normalization ensures smooth and bounded influence across training steps. The final quality weight is then given by:
\begin{equation}
w_t = w_{idx \bmod 7} \cdot (1 - \frac{\lambda \cdot f(\Delta Q_t)}{\lambda_{max}} + \frac{\lambda \cdot f(\Delta Q_{t-1})}{\lambda_{max}}),
\end{equation}
where $w$ defines 7 hierarchical quality weights adapted to Vimeo’s GOP setting. The modulation incorporates both current and previous frame quality variation through $\Delta Q_t$and $\Delta Q_{t-1}$. $\lambda$ is a hyperparameter set in the loss function during training to achieve different bitrates. Multiplying by $\frac{\lambda}{\lambda_{max}} $ helps ensure training stability across varying bitrate settings.

\subsection*{\textbf{Training-free Smooth Motion Estimation}}

\begin{figure}[t]
\centering
\includegraphics[width=0.8\textwidth]{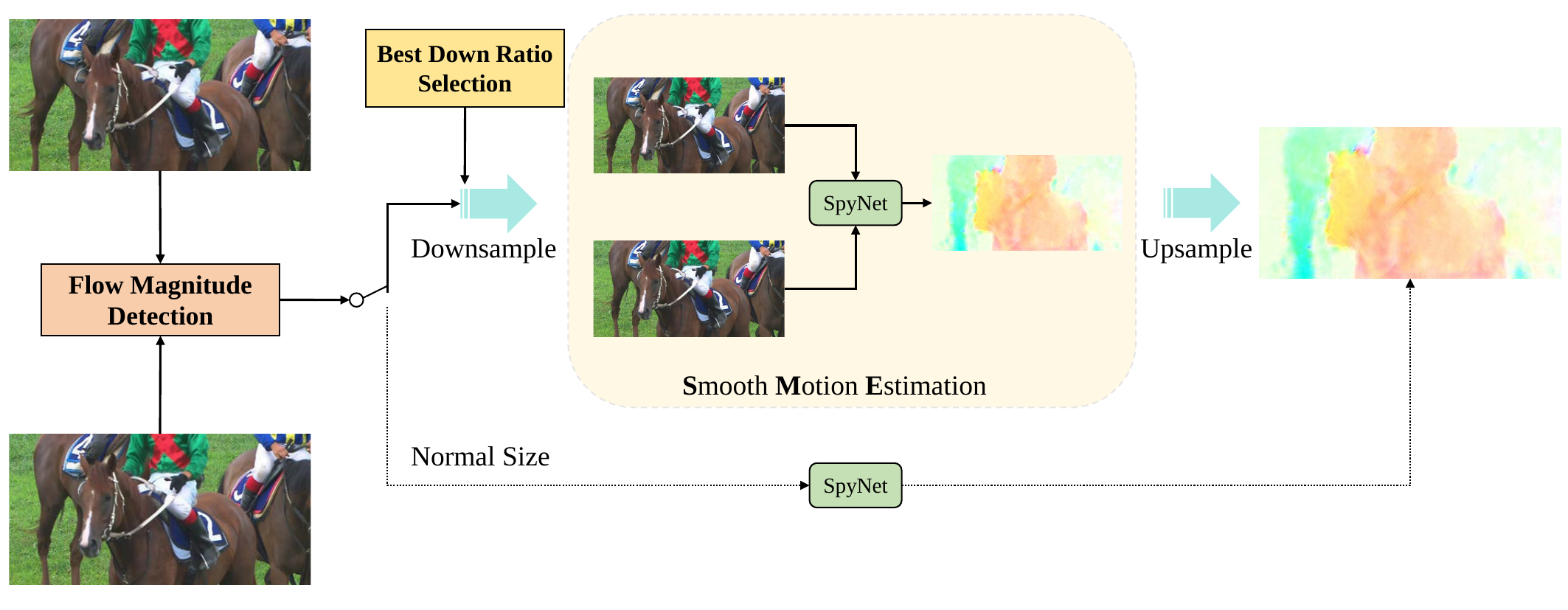}
\caption{The structure of the smooth motion estimation module.}
\label{figs:SME}
\end{figure}
To ensure robust flow accuracy in scenarios with large motion, we adopt the training-free Smooth Motion Estimation (SME) method. This approach was originally proposed by~\cite{bilican2025content} and subsequently proven effective in the AVS standard (M8967) for the EEM codec~\cite{sheng2025adap}. The key idea is to adaptively select the optimal downsampling scale for motion estimation by comparing alignment quality across different scales, as shown in Fig.~\ref{figs:SME}. By incorporating this strategy, our framework can robustly handle optical flow estimation for large motion and deformation without additional training costs.

\vspace{-0.5em}
\begin{algorithm}[H]
\resizebox{0.8\textwidth}{!}{
\captionsetup{font={small}} 
\caption{Content Adaptive Flow Estimation}
\label{alg:adaptive_flow}
\begin{minipage}{\textwidth}
\begin{algorithmic}[1]
\State \textbf{Input:} $x_t$, $\hat{x}_{t-1}$, scale set $\mathcal{D} = \{1, 1.25, 1.5, ..., 4.75, 5\}$
\State \textbf{Initialize:} $\mathrm{PSNR}_{\text{best}}= 0$, $D_{\text{best}}= 1$
\For{each $D_{\text{current}}$ in $\mathcal{D}$}
    \State Downsample $x_t$, $\hat{x}_{t-1}$ by $D_{\text{current}}$ to get $x_t'$, $\hat{x}_{t-1}'$
    \State Estimate $v_t'$ using SpyNet on $x_t'$ and $\hat{x}_{t-1}'$
    \State Upsample $v_t'$ to original resolution $v_t$
    \State Obtain $\tilde{x}_t$ through $ Warp( v_t,\hat{x}_{t-1})$
    \State Compute $\mathrm{PSNR}_{\text{current}}$ between $x_t$ and $\tilde{x}_t$
    \If{$\mathrm{PSNR}_{\text{current}} > \mathrm{PSNR}_{\text{best}} + \delta$}
        \State $\mathrm{PSNR}_{\text{best}}= \mathrm{PSNR}_{\text{current}}$
        \State $D_{\text{best}}= D_{\text{current}}$
    \EndIf
\EndFor
\State \Return $v_t$ estimated using $D_{\text{best}}$
\end{algorithmic}
\end{minipage}
}
\end{algorithm}
\vspace{-1em}

The module is activated when the average flow magnitude $\|{v}_t\|$ exceeds a threshold $\tau$:
\[
\|{v}_t\| = \sqrt{\frac{v_x^2 + v_y^2 + 10^{-8}}{H \times W}},
\]
where $v_x$ and $v_y$ are the horizontal and vertical components of the flow, $H \times W$ is the frame resolution, and $10^{-8}$ ensures numerical stability.

We define candidate scales $\mathcal{D} = \{1, 1.25, 1.5, ..., 4.75, 5\}$. For each scale $D_{\text{current}}$, we bilinearly downsample both the current frame $x_t$ and reference frame $\hat{x}_{t-1}$, estimating flow $v_t'$ with a pretrained SpyNet network. The flow is then upsampled to the original resolution, yielding $v_t$. Using $v_t$, we warp $\hat{x}_{t-1}$ to generate the predicted frame $\tilde{x}_t$. We evaluate motion estimation quality by computing the PSNR between $x_t$ and $\tilde{x}_t$, denoted as $\mathrm{PSNR}_{\text{current}}$. If $\mathrm{PSNR}_{\text{current}}$ exceeds the best score by a margin of $\delta = 0.1$, we update the best score and record the corresponding scale. After testing all scales, the one with the highest PSNR, $D_{\text{best}}$, is used for final flow estimation. This method improves motion alignment without additional training. In practice, $\tau$ and $\mathcal{D}$ serve as hyperparameters that can be optimized according to dataset characteristics. Specifically for the HEVC Class B dataset, we empirically set $\tau = 10$ and restrict the search space to $\mathcal{D} = \{1, 1.25\}$ to achieve a favorable trade-off between inference efficiency and coding performance.

\begin{figure*}
    \centering
    \includegraphics[width=\textwidth]{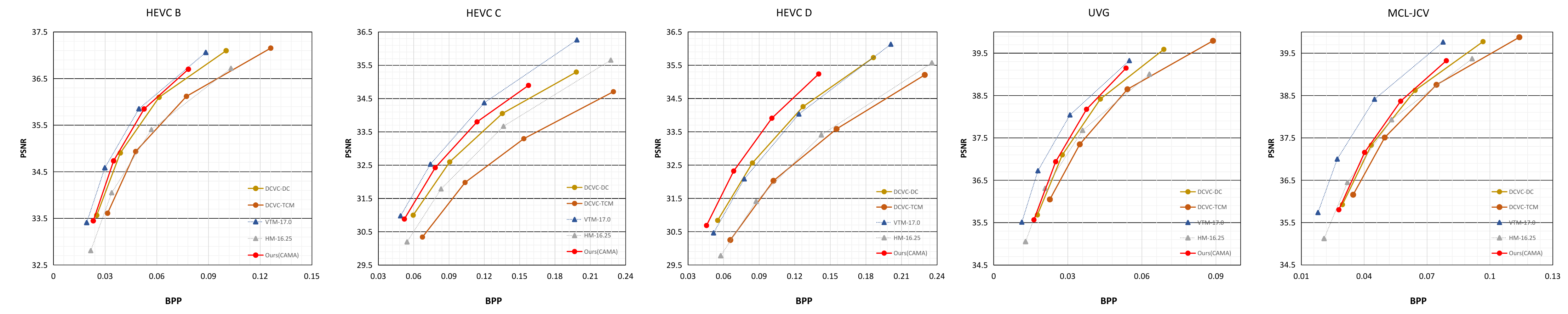}
    \caption{Rate-distortion curves on standard datasets, evaluated in RGB colorspace using PSNR for the first 96 frames, with the GOP size set to 32.}
    \label{fig:wide-top}

\hfill
\vspace{-0.5em}
\begin{minipage}{\textwidth}
    \centering
    \captionof{table}{BD-Rate (\%) comparison for PSNR and MS-SSIM on standard test datasets. The anchor is DCVC-TCM.}
    \label{tab:main-result}
    \resizebox{\textwidth}{!}{
    \begin{tabular}{>{\centering\arraybackslash}m{3cm}cccccc|cccccc}
    \toprule
    \multirow{2}{*}{\textbf{Method}} 
        & \multicolumn{6}{c}{\textbf{PSNR}} 
        & \multicolumn{6}{c}{\textbf{MS-SSIM}} \\
    \cmidrule(lr){2-7} \cmidrule(lr){8-13}
        & \textbf{HEVC B} & \textbf{HEVC C} & \textbf{HEVC D} & \textbf{UVG} & \textbf{MCL-JCV} & \textbf{AVG} 
        & \textbf{HEVC B} & \textbf{HEVC C} & \textbf{HEVC D} & \textbf{UVG} & \textbf{MCL-JCV} & \textbf{AVG} \\
    \midrule
    DCVC-TCM      & 0.00 & 0.00 & 0.00 & 0.00 & 0.00 & 0.00 & 0.00 & 0.00 & 0.00 & 0.00 & 0.00 & 0.00 \\
    HM-16.25      & -1.74 & -17.79 & -2.36 & -8.15 & -7.71 & -7.55 & 53.38 & 56.93 & 87.56 & 23.67 & 52.28 & 54.76 \\
    VTM-17.0      & -28.81 & -40.10 & -26.98 & -31.58 & -34.03 & -32.30 & 12.91 & 16.26 & 41.26 & 1.12 & 13.51 & 17.01 \\
    DCVC-DC       & -18.69 & -27.41 & -29.10 & -13.38 & -9.25 & -19.57 & -10.01 & -16.75 & -17.33 & -7.58 & -7.87 & -11.91 \\
    \textbf{Ours (CAMA)} & \textbf{-22.36} & \textbf{-34.33} & \textbf{-38.81} & \textbf{-17.64} & \textbf{-11.63} & \textbf{-24.95} & \textbf{-13.69} & \textbf{-21.06} & \textbf{-21.43} & \textbf{-10.60} & \textbf{-10.73} & \textbf{-15.50} \\
    \bottomrule
    \end{tabular}
    }
\end{minipage}
\vspace{-1em}
\end{figure*}

\Section{Experiments and Results}

\subsection*{\textbf{Datasets and Training Strategy}}

We adopt the Vimeo-90K~\cite{xue2019video} dataset for training, where video sequences are cropped into $256 \times 256$ patches. Our method is evaluated on the RGB colorspace of HEVC Class B, C, D~\cite{bossen2010common}, UVG~\cite{mercat2020uvg}, and MCL-JCV~\cite{wang2016mcl} test datasets, covering resolutions from $416{\times}240$ to $1920{\times}1080$. The training loss function balances distortion and bitrate via a rate-distortion trade-off:

\[
\mathcal{L} = w \cdot \lambda \cdot \mathcal{D} + \mathcal{R},
\]
where $\mathcal{D}$ is the distortion (MSE or 1-MS-SSIM), $\mathcal{R}$ is the bitrate, and $w$ is a hierarchical quality weight used only during finetuning.

The training process has two stages: pretraining and variable rate finetuning. Pretraining includes three phases: Motion, Context, and All. In the Motion phase, only motion-related modules are trained. In the Context phase, motion modules are frozen, and only context modules are optimized. In the All phase, all modules are trained jointly. During pretraining, $\lambda$ is fixed to 2048 for PSNR and 64 for MS-SSIM optimization. In the finetuning stage, we use four values for $\lambda$: $\{2048, 1024, 512, 256\}$ for PSNR and $\{64, 32, 16, 8\}$ for MS-SSIM, to train under different bitrate conditions.

\subsection*{\textbf{Experimental Result}}

We compare our model with VTM-17.0~\cite{vtm2022}, HM-16.25~\cite{hm2022}, DCVC-TCM, and our re-implementation of SOTA DCVC-DC under identical test conditions, using PSNR BD-rate (\%) as the evaluation metric. Negative values indicate bitrate savings. All models are evaluated on standard test datasets in RGB colorspace, with YUV sequences converted to RGB using the BT.709 standard. For VTM-17.0 and HM-16.25, YUV420 frames are first converted to RGB, then encoded and decoded in YUV444, and finally converted back to RGB for quality evaluation. We test the first 96 frames of each sequence with a GOP size of 32 and report the average results. The I-frame model follows the one in DCVC-DC. To meet input size requirements, all frames are padded to the nearest multiple of 16 before encoding, with bitrate calculated based on the original resolution. Actual bitstreams are written during encoding for realistic evaluation. As shown in Table~\ref{tab:main-result}, our method outperforms DCVC-DC and HM-16.25, with PSNR improving over DCVC-TCM by 24.95\% and MS-SSIM by 15.50\%, respectively. 

\begin{figure}[t!]
\centering
\includegraphics[width=0.8\textwidth]{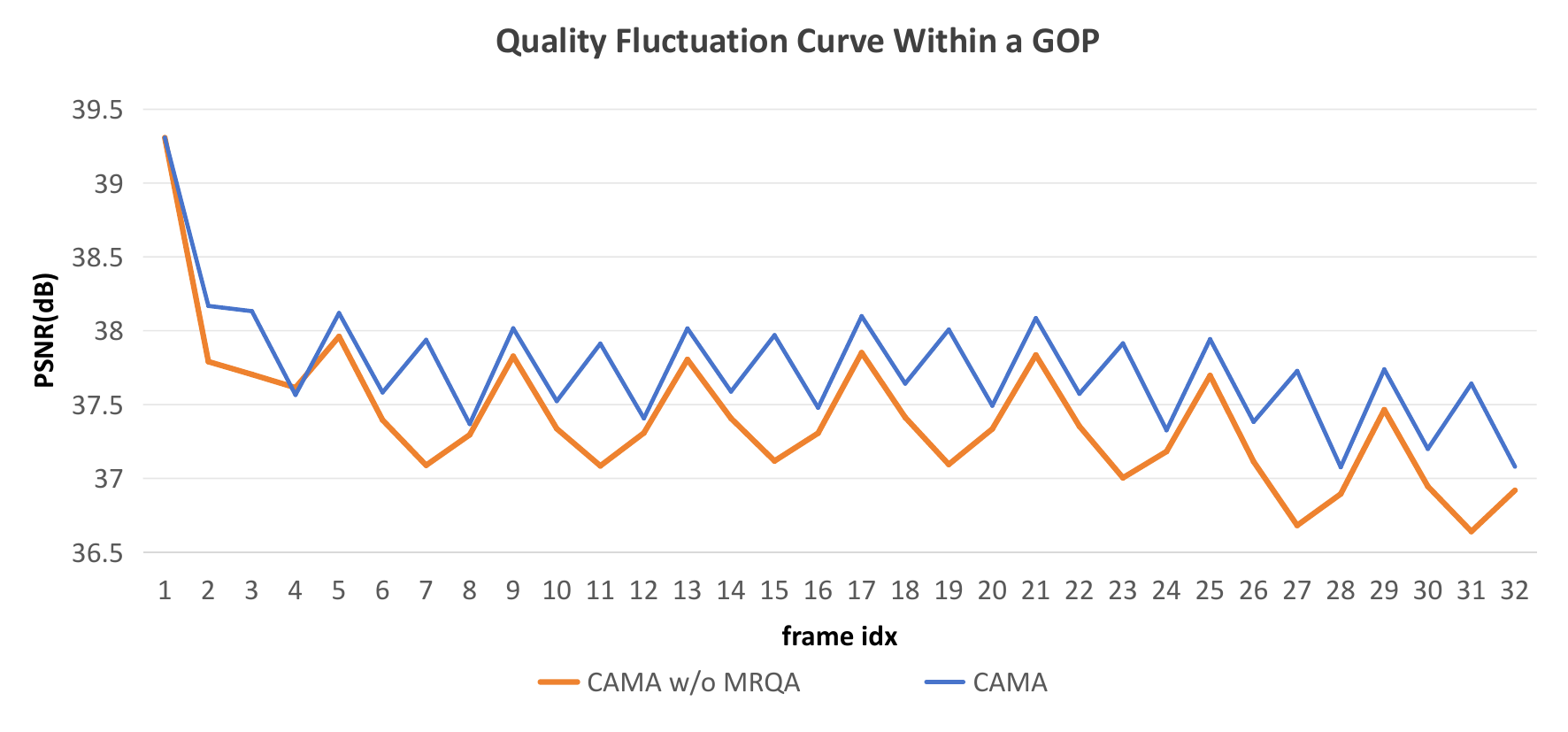}
\caption{Quality fluctuation within a GOP of HEVC B \textit{BasketballDrive} sequence.}
\label{figs:QualityFluctuation}
\vspace{-1em}
\end{figure}

We provide a subjective comparison in Fig.~\ref{figs:reconQuality}, showing frames from \textit{BasketballDrive}, \textit{Kimono}, and \textit{Cactus} in HEVC Class B, covering both dynamic and static scenes. Our method preserves finer details in regions with large motion or complex deformation. Fig.~\ref{figs:QualityFluctuation} illustrates the multi-reference hierarchical structure guided by MRQA, which improves quality consistency and reduces error propagation. Compared to prior approaches, our method produces smoother quality over time, while methods without MRQA exhibit larger fluctuations. As for complexity, only the TSMC module introduces new network components, while other changes affect training and inference computations. The parameter count increases by less than 5\%, and the decoding time on 1080p videos rises by no more than 5\% compared with DCVC-DC.

\begin{figure}[t!]
\centering
\includegraphics[width=0.8\textwidth]{figs/reconQuality.pdf}
\caption{Subjective comparison on frames from \textit{BasketballDrive}, \textit{Kimono}, and \textit{Cactus} in HEVC Class B. Our method achieves better quality with fewer bits.}
\label{figs:reconQuality}
\vspace{-1em}
\end{figure}

\subsection*{\textbf{Ablation Study}}
We perform ablation experiments on HEVC Class B to evaluate the individual and combined impact of the proposed modules, as summarized in Table~\ref{tab:ablation}. Introducing TSMC yields a notable BD-rate reduction of 2.83\% by enhancing motion alignment through lightweight deformable refinement. The MRQA module further improves performance with a 3.66\% BD-rate gain by dynamically adjusting hierarchical quality weights across frames based on content. Although SME alone brings a modest improvement of 0.31\%, it proves beneficial in sequences with large motion and high resolution. Combining TSMC and MRQA achieves a larger gain of 4.72\%, and integrating all three components results in the best overall performance, with a BD-rate reduction of 5.05\%.

\begin{table}[t!]
\centering
\caption{Ablation study for PSNR on HEVC Class B.}
\label{tab:ablation}
\renewcommand{\arraystretch}{1.1}
\resizebox{0.7\textwidth}{!}{%
\begin{tabularx}{\linewidth}{>{\centering\arraybackslash}c>{\centering\arraybackslash}X>{\centering\arraybackslash}X>{\centering\arraybackslash}X>{\centering\arraybackslash}X>
{\centering\arraybackslash}X>
{\centering\arraybackslash}X}
\toprule
\textbf{Method} & \textbf{Ma} & \textbf{Mb} & \textbf{Mc} & \textbf{Md} & \textbf{Me} & \textbf{Mf} \\
\midrule
\textbf{TSMC} & \makecell[c]{\xmark} & \makecell[c]{\cmark} & \makecell[c]{\xmark} & \makecell[c]{\xmark} & \makecell[c]{\cmark} & \makecell[c]{\cmark} \\
\textbf{MRQA} & \makecell[c]{\xmark} & \makecell[c]{\xmark} & \makecell[c]{\cmark} & \makecell[c]{\xmark} & \makecell[c]{\cmark} & \makecell[c]{\cmark} \\
\textbf{SME} & \makecell[c]{\xmark} & \makecell[c]{\xmark} & \makecell[c]{\xmark} & \makecell[c]{\cmark} & \makecell[c]{\xmark} & \makecell[c]{\cmark} \\
\midrule
\textbf{BD-rate (\%) ↓} & 0.00 & -2.83 & -3.66 & -0.31 & -4.72 & \textbf{-5.05} \\
\bottomrule
\end{tabularx}
}
\vspace{-1em}
\end{table}

\Section{Conclusion}
In this paper, we proposed a Content-Adaptive Motion Alignment (CAMA) framework to address the lack of content-specific adaptation in generalized learned video compression. Specifically, we first introduced a two-stage motion compensation module that utilizes flow-guided deformable alignment with coarse-to-fine offsets to achieve precise feature alignment. Second, we developed a multi-reference quality aware hierarchical training structure, which adaptively adjusts distortion weights based on reference quality fluctuations to effectively reduce error propagation. Third, we integrated a training-free smooth motion estimation module that performs adaptive downsampling based on motion magnitude to handle complex motion. Experimental results demonstrate that CAMA achieves significant performance gains, yielding a 24.95\% BD-rate reduction over the baseline DCVC-TCM and outperforming both the state-of-the-art DCVC-DC and the traditional HM-16.25 codec.

\Section{References}
\bibliographystyle{ieeebib}
\bibliography{refs}

\end{document}